\documentclass[runningheads]{llncs}
\usepackage{graphicx}
\usepackage{xcolor}
\usepackage{multirow}
\usepackage{float}

\usepackage{color}
\usepackage{colortbl}
\definecolor{LightGray}{rgb}{0.9,0.9,0.9}
\begin{document}
\title{Observation Denoising in CYRUS Soccer Simulation 2D Team For RoboCup 2023}
%
%
\author{
Aref Sayareh\inst{3}\and
Nader Zare\inst{1}\and 
Omid Amini\inst{1}\and
Arad Firouzkouhi\inst{4}\and
Mahtab Sarvmaili\inst{1}\and
Stan Matwin\inst{1,}\inst{2} \and
Amilcar Soares\inst{3}}
\authorrunning{Aref et al.}
%
\institute{
Institute for Big Data Analytics, Dalhousie University, Halifax, Canada\\
\and
Institute for Computer Science, Polish Academy of Sciences, Warsaw, Poland\\
\and
Memorial University of Newfoundland, St. John's, Canada\\
\and
Amirkabir University of Technology, Iran\\
\email{nader@cyrus2d.com},\\ 
\email{\{omid.amini, mahtab.sarvmaili\}@dal.ca},\\ \email{arad.firouzkouhi@aut.ac.ir}\\ 
\email{stan@cs.dal.ca}, \\
\email{\{asayareh, amilcarsj\}@mun.ca}
}
\maketitle              
\begin{abstract}
The RoboCup competitions hold various leagues, and the Soccer Simulation 2D League is a major one among them. Soccer Simulation 2D (SS2D) match involves two teams, including 11 players and a coach, competing against each other. The players can only communicate with the Soccer Simulation Server during the game. 
This paper presents the latest research of the CYRUS soccer simulation 2D team, the champion of RoboCup 2021.
We will explain our denoising idea powered by long short-term memory networks (LSTM) and deep neural networks (DNN). The CYRUS team uses the CYRUS2D base code that was developed based on the Helios and Gliders bases.

\keywords{Robotic \and Denoising \and Soccer Simulation.}
\end{abstract}
\section{Introduction}
RoboCup has hosted annual robotic soccer competitions since 1997, following the proposal of robotic soccer games as a new research topic in 1992\cite{robo1997,noda1996soccer,kitano1997robocup}. The primary objective of these competitions is to advance the robotics and artificial intelligence field by implementing robotic soccer games. Several leagues, including rescue, soccer simulation, and standard platform soccer, have been created to achieve this.

The Soccer Simulation 2D (SS2D) league is one of the oldest RoboCup leagues. In a soccer simulation 2D game, two teams containing 11 players and one coach compete against each other. Each game takes 6000 cycles (10 minutes). In each cycle, the players and coach receive observations from the RoboCup Soccer Simulation Server and send actions based on those observations back to the server. The server is responsible for controlling the game.

CYRUS has been a regular participant in RoboCup competitions since 2013 and emerged as the champion of the SS2D league in RoboCup 2021. Over the years, the CYRUS team has achieved first, second, second, third, fourth, and fifth places in RoboCup in 2021, 2022, 2018, 2019, 2017, and 2014, respectively. CYRUS also achieved first place in IranOpen in 2018 and 2014, RoboCup Asia-Pacific in 2018, and second place in JapanOpen in 2020. 

In this paper, we will first discuss related research by other teams, focusing on topics such as AI and robotics in soccer simulation 2D. We will then present our previous works, including our released applications and codes. Next, we will explore the issue of noise in soccer simulation 2D, presenting our ideas for addressing this challenge. Finally, we will outline our plans for future work in this area.

\subsection{Related Work}

In this part, we will mention some latest novel ideas presented by SS2D teams in recent years.

YuShan2022 adopted an offline training model and LambdaMART algorithm to enhance their team's offensive ability through movement direction selection\cite{yushan22}.
RobôCIn researched an adaptive strategy based on time, score, and the opponent team's pressure. Also, they developed and released a log analyzer software\cite{robocin22}.
Persepolis team improved main behaviours such as pass, shoot, dribbling, and marking; also used a ranking algorithm to improve chain-action performance\cite{pers22}.
Persepolis also implemented a Path Planning algorithm to improve its offensive strategy\cite{pers21,razi19}.
Helios2022 conducted numerical experiments to investigate the effect of using Anonymous mode on game results\cite{hel22}.
HfutEngine2D used genetic algorithms to adjust parameters and attempted to use neural networks for rational decision-making, significantly improving the team's strength during recent tests against other RoboCup teams.\cite{hfut22}.
Alice2D's work centered around optimizing defense strategy using the Monte-Carlo tree and reinforcement learning algorithms.\cite{alice22}.
Hades2D team improved individual players' offensive and defensive skills, significantly decreasing the average number of conceded goals against the base.\cite{hades21,hades22}.
FRA-UNIted utilized reinforcement learning (RL) techniques to improve their ability to intercept the ball from a player in possession. They created a Markov decision process (MDP) environment with two players: the agent learning to intercept and the dribbling player. Through this environment, the agent learned and improved its intercepting skills over time.\cite{fra22}.

\subsection{CYRUS Previous Works}
 In this section, we briefly explain the history of our research. 
CYRUS was established in 2013 and has participated in many competitions since then. CYRUS Team researched various aspects of the game, starting with passing behavior, unmarking strategy, and formation\cite{cyrus13,cyrus14}. After that, we shifted our attention to defensive behaviour\cite{cyrus15}, shooting algorithms, and communication modules\cite{cyrus16}. In 2017, the CYRUS team developed the opponent behavior prediction module\cite{cyrus17,cyrus18}. In recent years, CYRUS Team has mainly been developing the pass prediction and unmarking module, and Data Generator framework \cite{cyrus19,cyrus21,cyrus22,cyrussamposiom,cyruschamp}. Some of our research work, including Cyrus2D Base code, CYRUS 2014 Source Code, PYRUS - Python SS2D Base Code, and SS2D Log Analyzer, has been released in recent years.

\subsubsection{Cyrus 2014} Cyrus team released its source code used in the 2014 RoboCup competition, where they achieved fifth place. The source code can be an excellent resource for other teams and new teams interested in participating in the RoboCup competition. The code provides valuable insights into the team's approach, strategies, and implementations that can be helpful for others to improve their performance. The code is available on our GitHub\footnote{Cyrus 2014 Source \url{https://github.com/naderzare/cyrus2014}.}.

\subsubsection{Cyrus2D Base} Cyrus2D Base\cite{cyrusbase} code is a soccer simulation base code created by combining Helios Base\cite{heliosbase}, Gliders base\cite{glbase}, and the best features of the Cyrus team, which won first place in the RoboCup Soccer Simulation 2D league in 2021.  This base code is an excellent starting point for new teams just starting to work on their soccer simulation projects. The code provides a solid foundation for building the functionalities required for the RoboCup competition. The code is available on our GitHub\footnote{Cyrus Base Source \url{https://github.com/Cyrus2D/Cyrus2DBase}}.

\subsubsection{PYRUS Base} Pyrus Base code is a soccer simulation base code developed in Python, providing an excellent platform for researchers to develop and test their Machine Learning algorithms. This base code benefits teams interested in applying artificial intelligence techniques to enhance performance. The code is available on our GitHub\footnote{Pyrus Base \url{https://github.com/Cyrus2D/Pyrus-SS2D-Base}}.

\subsubsection{Cyrus AutoTest2D} This software was initially developed by the WrightEagle team\cite{wrighteagle}, and Cyrus team members upgraded and simplified it. The software helps teams run many parallel games to test their performance. Cyrus AutoTest provides a systematic approach to evaluating teams' performance in the soccer simulation league, and the software can help teams identify their strengths and weaknesses to improve their performance. The code is available on our GitHub\footnote{AutoTest \url{https://github.com/Cyrus2D/AutoTest2D}}.

\subsubsection{AutoTune2D} Cyrus team developed this tool that allows researchers to fine-tune their parameters alongside AutoTest. AutoTune can be a valuable tool for researchers who want to optimize their team's performance by adjusting various parameters to find the best combination. The code is available on our GitHub\footnote{AutoTune \url{https://github.com/Cyrus2D/AutoTune2D}}.

\subsubsection{Cyrus Log analyzer} This software can analyze soccer simulation 2D game logs to find statistical values, including the number of passes, dribbles, and other relevant statistics. This software can provide valuable insights into a team's performance, helping teams identify areas of improvement and fine-tune their strategies. The Cyrus Log analyzer can help teams optimize their performance by providing detailed information about their gameplay. The code is available on our GitHub\footnote{CyrusLog Analayzer\url{https://github.com/Cyrus2D/CyrusLogAnalyzer}}.

\section{Denoising}
\subsection{Observation Noise In SS2D}
In the Soccer Simulation 2D league, accurately calculating the position of each object, including players and the ball, presents a significant challenge due to agents' visual sensor limitations.





In this multi-agent environment, the agent can use three different vision angles: 180(Wide), 120(Normal), and 60(Narrow) degrees. However, using each angle comes at a cost. For instance, if the agent chooses to use the Wide angle, it must wait three cycles before using the sensor again. Similarly, for the Normal angle and Narrow angle, it should wait for 2 and 1 cycle, respectively. 
Choosing the right vision angle is important to balance performance and reduce the delay caused by limited sensor usage. 

To address this issue, a parameter called 'pos\_count' is defined for each object, which indicates the number of cycles that have elapsed since the last time the sensor received the object's position.

In addition, the relative distance and angle information that the agent receives from the visual sensor are quantized, resulting in more noise in the distance measurement when the object is farther away. Furthermore, the floating-point part of the angle measurement is rounded.

\subsection{Helios Base Denoising Method}
HeliosBase applies a technique to tackle the quantized noise by considering the object's distance. Specifically, it calculates the average range of distances at which the object can be detected. This method is also applied to the relative angle between the player and the object to refine the estimation further. After these initial estimates, the method can determine the object's relative position before calculating its global position. This approach allows for improved accuracy in object localization, which is crucial in many applications such as robotics and computer vision.

\subsection{Denoising by Machine Learning.}
Our proposed approach aims to reduce the noise in the data when the object's pos\_count is more than a cycle. To achieve this, we plan to investigate various machine learning techniques, including Deep Neural Networks (DNNs)\cite{dnn} and Long Short-Term Memory (LSTM)\cite{lstm} networks. These models' predicted positions are closer to the true objects' than the last seen positions. The time series models will be trained on input sequences that include the position and sensor data from the last five cycles. We aim to find the best machine learning model to predict the opponents' position.

\subsection{Generating Data}
In order to gather the necessary data for our study, we ran a total of 400 games between Cyrus2DBase and HeliosBase. The collected data includes the positional and velocity information of the ball and 22 players within the game and the direction of the players' bodies. The data also includes the pos\_count parameter for each player. The position data of each object consists of both noisy and original values.






\section{Results}
In addition to training the DNN and LSTM models, we also experimented with different sizes and configurations of these models to find the optimal architecture for denoising the positional data. After extensive testing, we found the LSTM model with 512 and 256 neurons and all ReLU activations in the time series model and the DNN model with 512, 256, 128, 64, and 32 neurons all ReLU activations achieved the best performance.
Furthermore, we discovered that the LSTM model was more effective in predicting the accurate position of the players than the DNN model.
Therefore, we focused on improving our solution's DNN and LSTM models.

\begin{figure}[]
    \centering
    \includegraphics[width=0.8\textwidth]{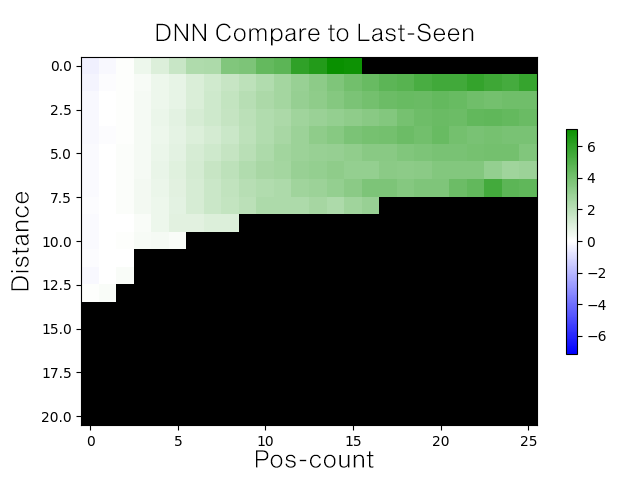} \\
    \caption{Comparison of DNN accuracy and last-seen position accuracy. The green color shows a better performance with DNN, whereas the blue color indicates better performance in the last seen position. Insufficient data to compare these two methods are shown in black color.}
    \label{fig:label1}
\end{figure}
\begin{figure}[]
    \centering
    \includegraphics[width=0.8\textwidth]{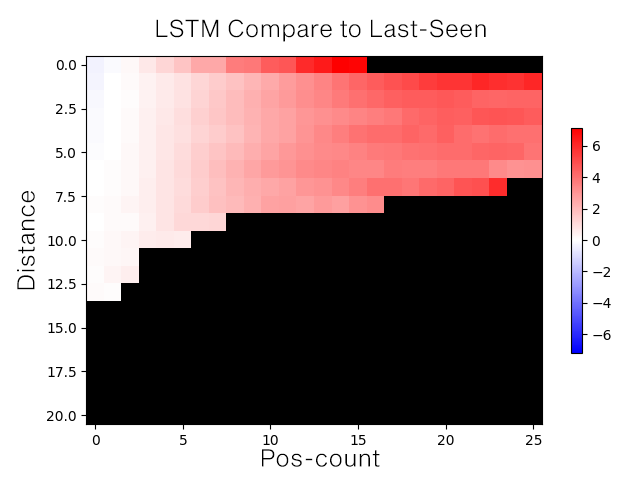} \\
    \caption{Comparison of LSTM accuracy and last-seen position accuracy. The red color shows a better performance with LSTM, whereas the blue color indicates better performance in the last seen position. Insufficient data to compare these two methods are shown in black color.}
    \label{fig:label2}
\end{figure}
\begin{figure}[]
    \centering
    \includegraphics[width=0.8\textwidth]{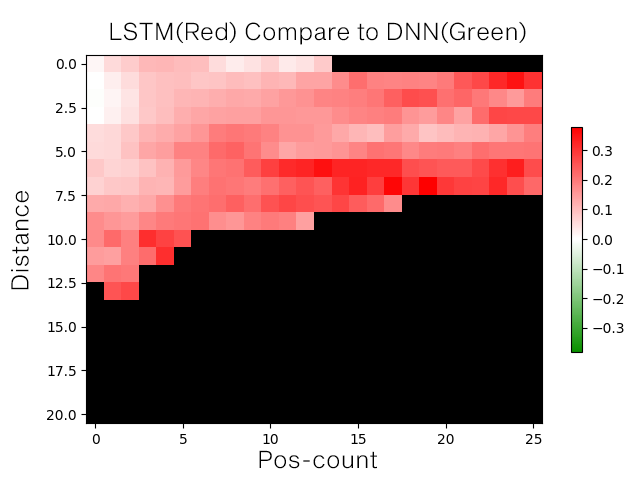} \\
    \caption{Comparison of LSTM accuracy and DNN accuracy. The RED color shows a better performance with LSTM, whereas the green color indicates better performance in DNN. Insufficient data to compare these two methods are shown in black color.}
    \label{fig:label3}
\end{figure}

Figures \ref{fig:label1}, \ref{fig:label2} and \ref{fig:label3} presented in the heatmap compare errors among three methods: DNN, LSTM, and Agent Vision data. The heatmaps are based on the number of cycles an agent has not seen the player and the distance between the agent and the player. The agent represented in the figures is number 9 of the left team, and the object being observed is player 5 of the left team. LSTM has a higher level of accuracy than DNN, while DNN is more precise than player vision. Therefore, LSTM performs more accurately than the other two methods for this specific task.

\section{Future work}
There are several challenges associated with the proposed models. Firstly, the accuracy of the models may decrease if a team's strategy and formation change, as the models primarily learn from the team's historical data. Secondly, there may not be existing models for new teams in the league due to the lack of data. In such cases, one approach could be choosing a pre-trained model for other teams similar to the new ones to predict their performance. Another approach could be implementing online learning techniques that learn the opponent's strategy during a game, which can be challenging and require further investigation.

\end{document}